\documentclass{article}
\usepackage{spconf,amsmath,graphicx}

\usepackage{graphicx}
\usepackage{xr}
\usepackage{algpseudocode}
\usepackage{algorithm}
\usepackage{hyperref}
\usepackage[capitalize,nameinlink]{cleveref}
\let\OLDthebibliography\thebibliography
\renewcommand\thebibliography[1]{
  \OLDthebibliography{#1}
  \setlength{\parskip}{0pt}
  \setlength{\itemsep}{0pt plus 0ex}
}
\usepackage{appendix}
\usepackage{empheq}
\usepackage{enumitem}
\usepackage{array}
\usepackage{color}
\definecolor{umn_maroon}{RGB}{122, 0, 25}
\usepackage[numbers,sort&compress]{natbib}

\title{Practical Phase Retrieval Using Double Deep Image Priors}
\name{Zhong Zhuang$^1$,
        David Yang$^2$,
        Felix Hofmann$^2$,
        David A. Barmherzig$^3$, 
        Ju Sun$^4$}
\address{$^1$Electrical \& Computer Engineering, University of Minnesota, Minneapolis, USA\\
        $^2$Department of Engineering Science, University of Oxford, Oxford, UK \\
        $^3$Center for Computational Mathematics, Flatiron Institute, New York City, USA \\
        $^4$Computer Science \& Engineering, University of Minnesota, Minneapolis, USA}
\usepackage{graphicx,amsmath,amsfonts,amscd,amssymb,bm,url,color,latexsym}

\allowdisplaybreaks

\renewcommand{\mathbf}{\boldsymbol}

\newcommand{\mb}{\mathbf}
\newcommand{\mc}{\mathcal}

\newcommand{\bb}{\mathbb}

\newcommand{\set}[1]{\left\{ #1 \right\}}

\newcommand{\reals}{\bb R}

\newcommand{\eps}{\varepsilon}
\newcommand{\R}{\reals}
\newcommand{\Cp}{\bb C}

\makeatletter
\newcommand*{\addFileDependency}[1]{
  \typeout{(#1)}
  \@addtofilelist{#1}
  \IfFileExists{#1}{}{\typeout{No file #1.}}
}
\makeatother

\newcommand{ \brac }[1]{\left[ #1 \right]}
\newcommand{ \paren }[1]{ \left( #1 \right) }

\DeclareMathOperator{\st}{s.t.}

\newcommand{\wh}{\widehat}

\newcommand{\T}{\intercal}

\newcommand{\abs}[1]{\left| #1 \right|}
\newcommand{\innerprod}[2]{\left\langle #1,  #2 \right\rangle}

\numberwithin{equation}{section}

\begin{document}
%
\maketitle

\begin{abstract}
Phase retrieval (PR) concerns the recovery of complex phases from complex magnitudes. We identify the connection between the difficulty level and the number and variety of symmetries in PR problems. We focus on the most difficult far-field PR (FFPR), and propose a novel method using double deep image priors. In realistic evaluation, our method outperforms all competing methods by large margins. As a single-instance method, our method requires no training data and minimal hyperparameter tuning, and hence enjoys good practicality. 

\end{abstract}
\section{Introduction}
\label{sec:intro}
In scientific imaging, observable physical quantities about the object of interest are often complex-valued, e.g., when diffraction happens~\cite{Goodman2017Introduction}. However, practical detectors can only record complex magnitudes, but not phases, resulting in phaseless observations. Phase retrieval (PR), broadly defined, is the nonlinear inverse problem (NIP) of estimating the object of interest from the phaseless observations. PR is central to coherent diffraction imaging ((B)CDI)~\cite{miao1999extending,RobinsonEtAl2001Reconstruction}, image-based wavefront sensing~\cite{LukeEtAl2002Optical}, radar and sonar sensing~\cite{Jaming1999Phase}; see the recent survey~\cite{ShechtmanEtAl2015Phase}. 

\vspace{0.5em}\noindent\textbf{\textcolor{umn_maroon}{Which phase retrieval (PR)?}} \quad 
Without loss of generality, consider a 2D object of interest $\mb X \in \Cp^{m \times n}$, and a physical observation model $\mc A$ that leads to an ideal complex-valued observation $\mc A(\mb X) \in \Cp^{m' \times n'}$\footnote{3D phase retrieval problems can be formalized similarly. }. However, the detector can only record $\mb Y = \abs{\mc A(\mb X)}^2$, where $\abs{\cdot}^2$ denotes the elementwise squared magnitudes---which correspond to the photon flux the detector is able to capture. In \textbf{far-field (Fraunhofer) PR (FFPR) that stems from far-field propagation and is also the focus of this paper}, $\mc A$ is the oversampled 2D Fourier transform $\mc F$ with $m' \ge 2m-1$ and $n' \ge 2n-1$ to ensure recoverability. For notational simplicity, we assume $m, n$ are odd numbers, and $\mb X$ is centered at $(0, 0)$ so that index ranges are $[-(m-1)/2, (m-1)/2]$ and $[-(n-1)/2, (n-1)/2]$, respectively. 
Numerous other $\mc A$'s have been studied in the literature, notable ones including: 
\begin{itemize}[leftmargin=0.15in,itemsep=0.1em] 
    \item \textbf{Generalized PR (GPR)}: $\mc A(\mb X) = \set{\innerprod{\mb A_i}{\mb X}}_{i=1}^k$ where $\mb A_i$'s are iid Gaussian or randomly-masked Fourier basis matrices~\cite{candes2015phase,FannjiangStrohmer2020numerics}. These elegant mathematical models do not correspond to physically feasible imaging systems so far; 
    \item \textbf{Near-Field (Fresnel) PR (NFPR)}: $\mc A(\mb X) = \mc F(\mb X \odot [e^{i\pi \beta \paren{i^2 + j^2}}]_{i, j})$ \cite{WangEtAl2020Phase,ZhangEtAl2021PhaseGAN}, where the constant $\beta > 0$ depends on the sampling intervals, wavelength, and imaging distance~\cite{ShimobabaIto2019Computera}, comes from near field propagation. Note that FFPR corresponds to $\beta \to 0$, and PR problems solved in image-based wavefront sensing for astronomical applications correspond to multi-plane near-field propagation with sequential optical aberrations~\cite{Fienup1993Phase};  
    \item \textbf{Holographical PR (HPR)}: $\mc A(\mb X) = \mc F\paren{[\mb X, \mb R]}$, where $\mb R$ is a known reference that is put side-by-side with the object of interest $\mb X$~\cite{BarmherzigSun2022Towards}; depending on the propagation distance, near-field versions are also possible~\cite[Chapter 11]{Goodman2017Introduction}; 
    \item \textbf{Ptychography (PTY)}: $\mb X$ is raster scanned by a sharp illumination pattern $\mb W$ that is focused over a local patch of $\mb X$ each time. Now $\mb Y$ is the set of magnitude measurements $\mb Y_i = \abs{\mc F\paren{\mb W \odot \mb X(\mb p_i)}}^2$, where $\mb p_i$ indexes the raster grid~\cite{ThibaultEtAl2008High,MarchesiniEtAl2013Augmented}. 
\end{itemize} 

\begin{figure}[!htbp]
    \centering
    \vspace{-1em}
    \includegraphics[width=0.9\linewidth]{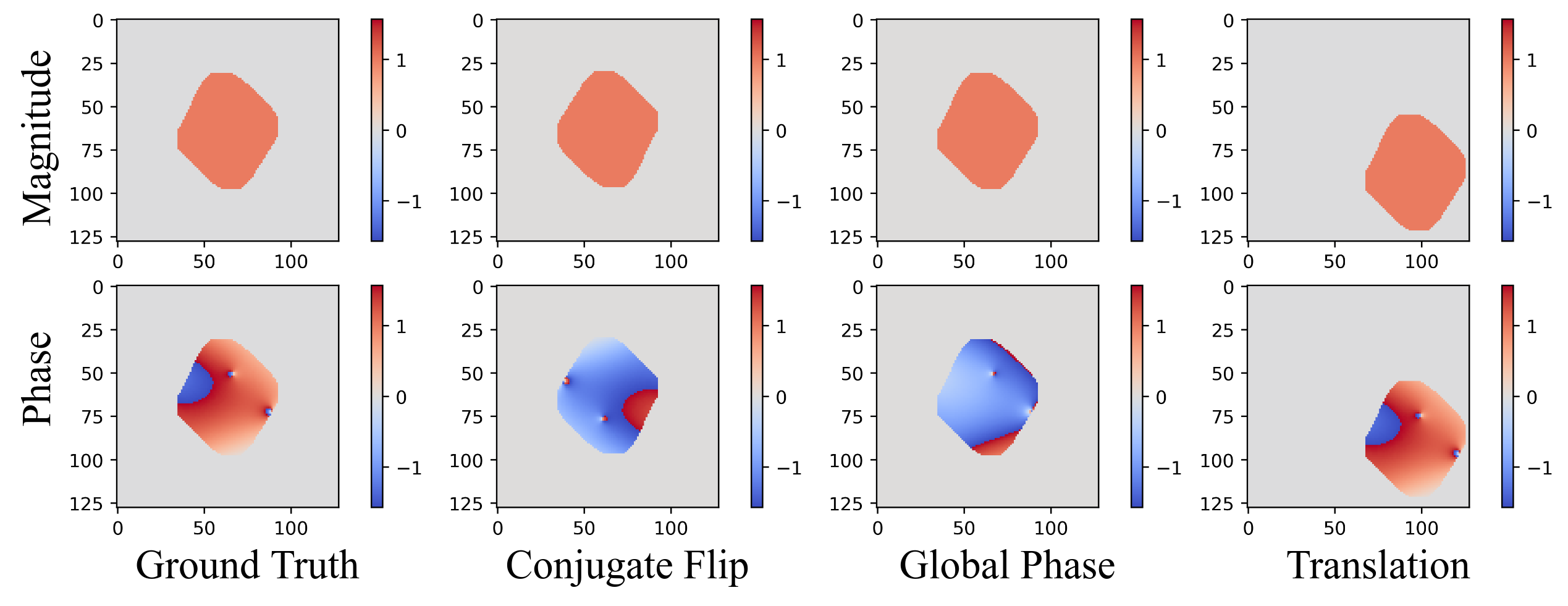}
    \vspace{-1em}
    \caption{Illustration of the three intrinsic symmetries in FFPR on simulated complex-valued crystal data (see \cref{sec:exps} for details). Any composition of 2D conjugate flipping, translation, and global phase, when applied to $\mb X$, leads to the same set of magnitudes $\mb Y$. }
    \label{fig:BCDI_sym}
\end{figure}
\begin{table*}[t]
    \centering
    \caption{Comparison of GPR, NFPR, and FFPR in terms of their symmetries and numerical solvability with the least-squares formulation in \cref{eq:LS4PR} combined with gradient descent.}
    \vspace{0.5em}
    \label{tab:symm_comp}
    \begin{tabular}{m{0.1\textwidth} | m{0.22\textwidth} | m{0.22\textwidth} | m{0.22\textwidth}}
    \hline \hline 
    PR model & GPR & NFPR &  FFPR \\
    \hline
   Symmetry & global phase & global phase & global phase, translation, 2D conjugate flipping  \\
   \hline 
   Final loss of solving LS in \cref{eq:LS4PR} using gradient descent from $100$ random initializations & \includegraphics[width=0.22\textwidth]{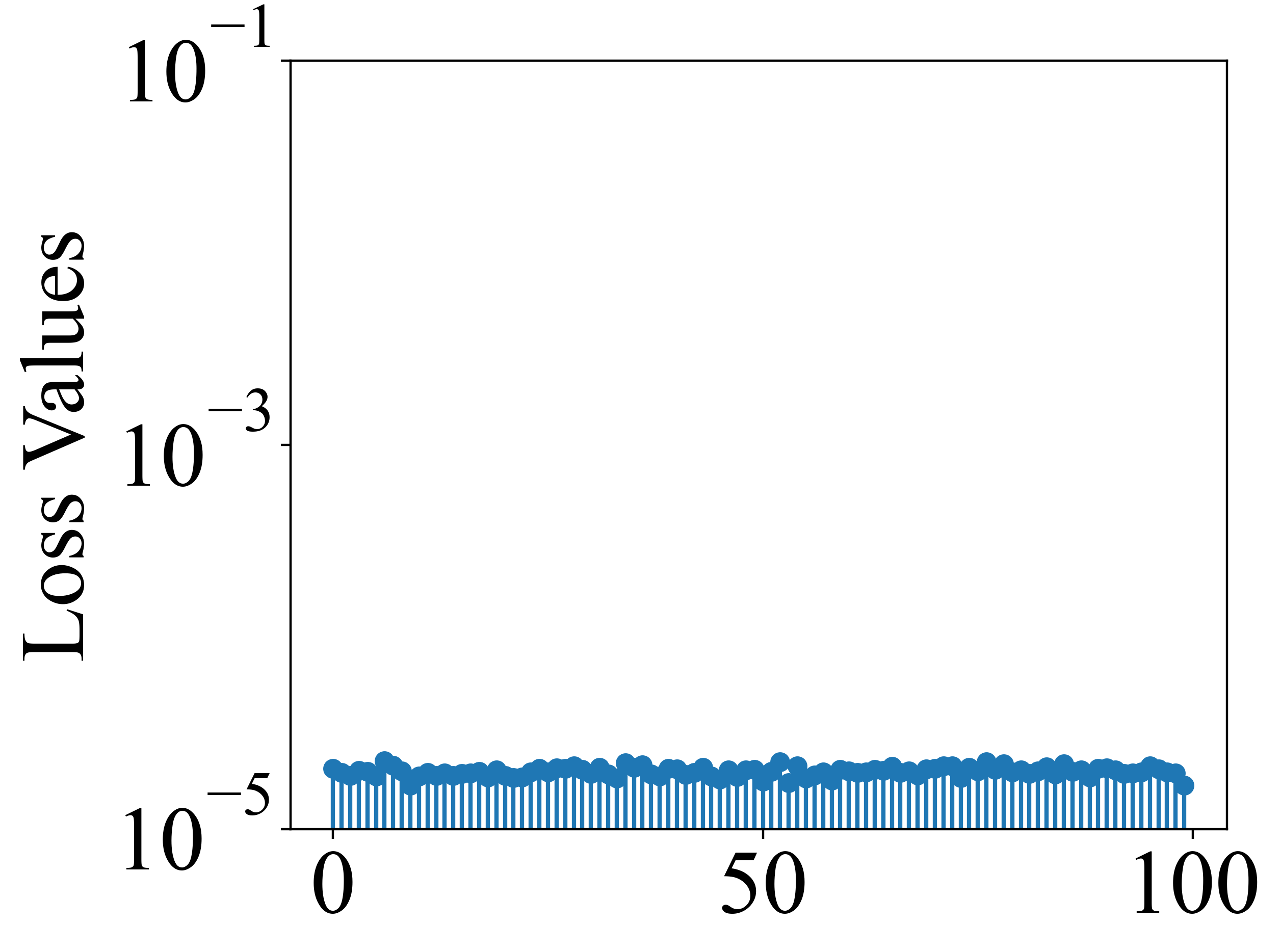}  & \includegraphics[width=0.22\textwidth]{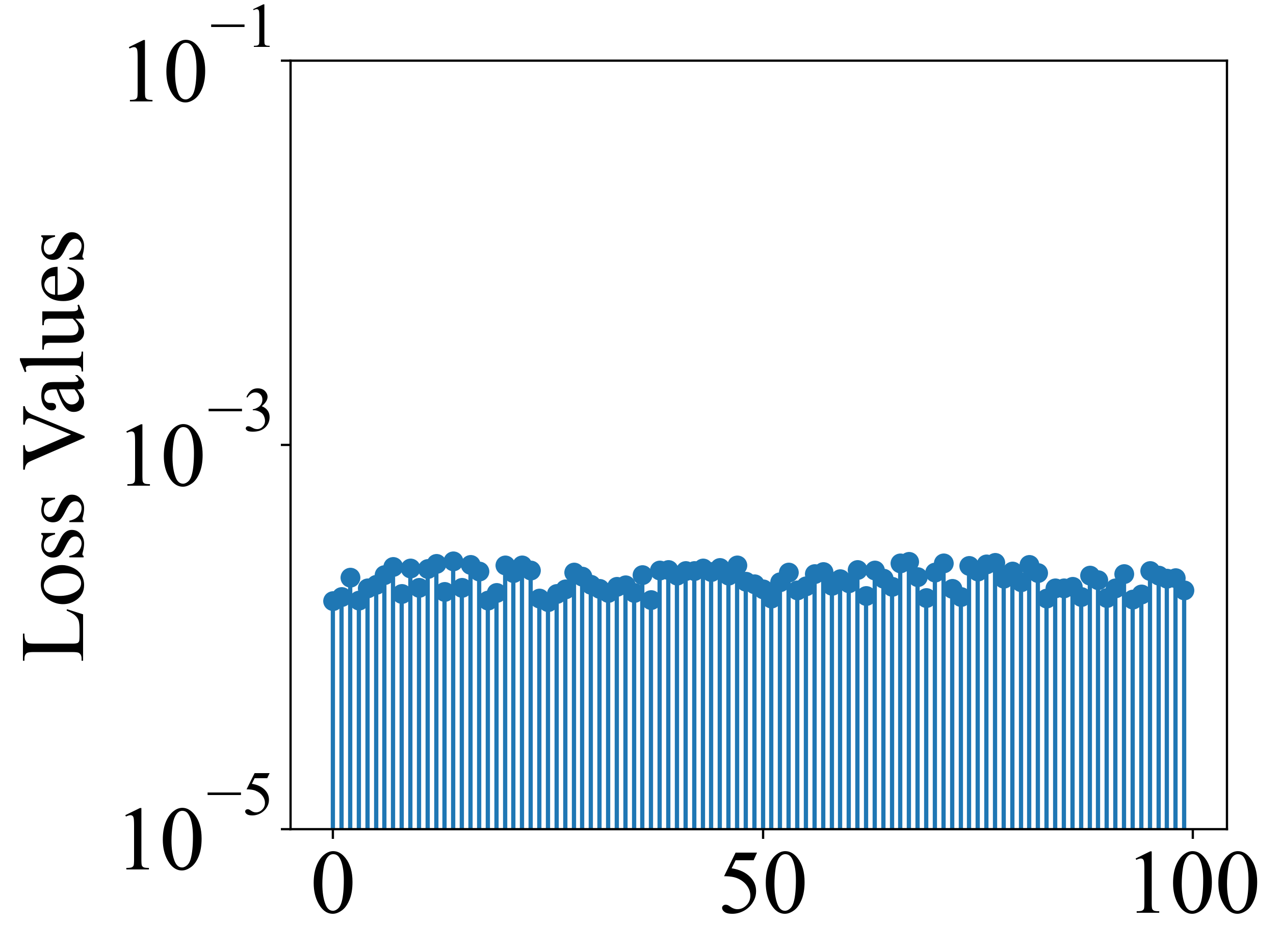} &  \includegraphics[width=0.22\textwidth]{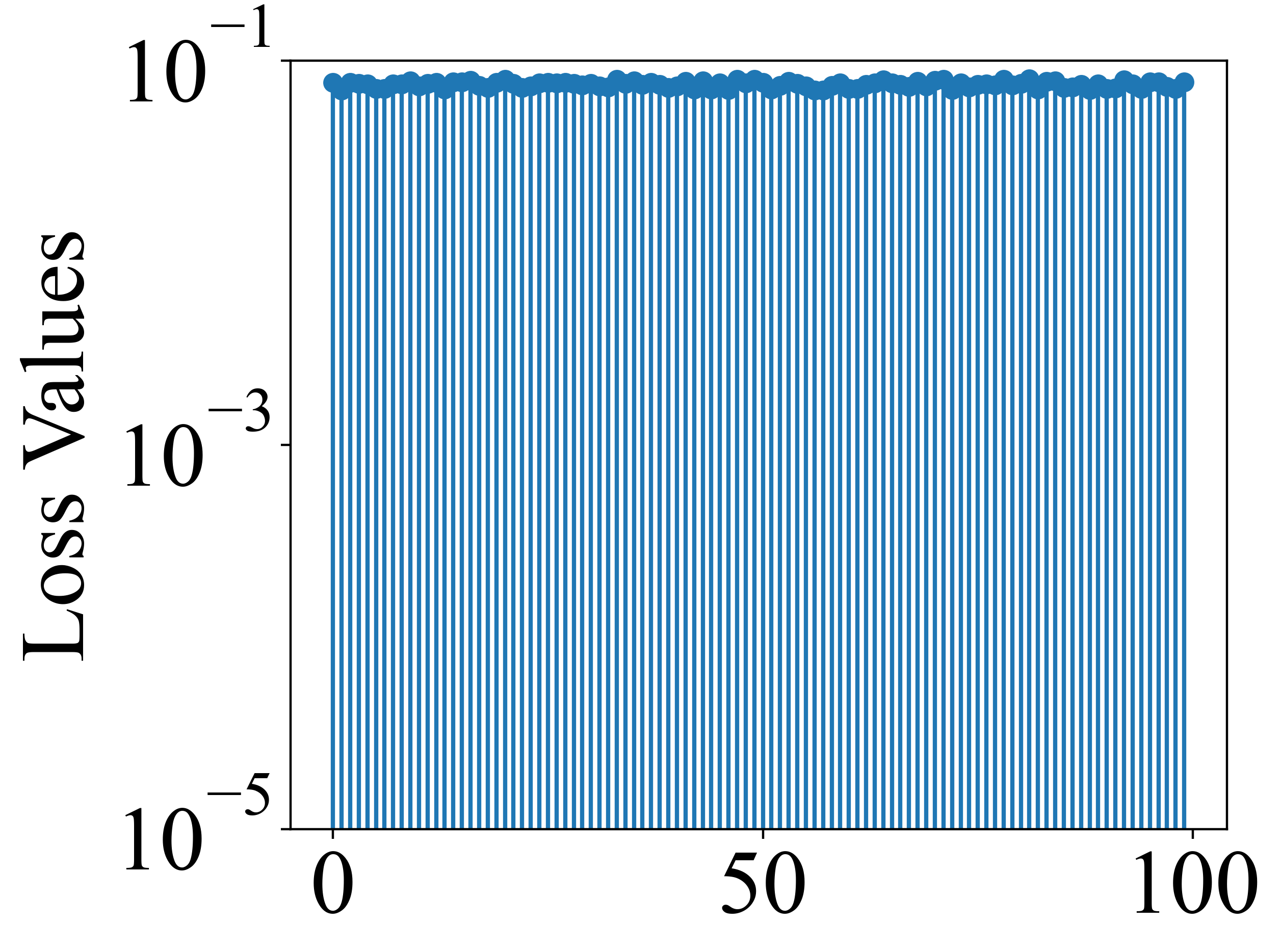} \\
   \hline \hline 
    \end{tabular}
\end{table*}
\noindent\textbf{\textcolor{umn_maroon}{Symmetry matters}} \quad 
Identifiability in PR is often up to intrinsic symmetries. For example, any global phase factor $e^{i \theta}$ added to $\mb X$ leaves $\mb Y$ unchanged for FFPR, NFPR, GPR, and PTY, i.e., \textbf{global phase symmetry}. While this is the only symmetry for NFPR, GPR, and PTY, FFPR has two other symmetries: translation and 2D conjugate flipping, as demonstrated in \cref{fig:BCDI_sym}~\cite{BendoryEtAl2017Fourier}. 
A crucial empirical observation is that \textbf{the difficulty level of a PR problem is proportional to the number of its symmetries}. To see the point, consider a natural least-squares (LS) formulation of PR\footnote{We consider this inner-unsquared version as it has better noise stability than the inner-squared one in practice~\cite{yeh2015experimental}. } 
\begin{align}  \label{eq:LS4PR}
    \min_{\mb X \in \Cp^{m \times n}} \; \frac{1}{m'n'}\|\sqrt{\mb Y} - \abs{\mc A(\mb X)}\|_F^2,  
\end{align} 
with the groundtruth complex-valued 2D crystal sample in \cref{fig:BCDI_sym} as the target $\mb X$.  On GPR with Gaussian, NFPR, and FFPR, we run gradient descent with $100$ random starts respectively and record their final convergent losses. As evident from \cref{tab:symm_comp}, while we can consistently find numerically satisfactory solutions for GPR and NFPR, we always find bad local solutions for FFPR---which has three symmetries. Similarly, for FFPR, 
\begin{figure}[!htbp]
    \centering
    \includegraphics[width=0.85\linewidth]{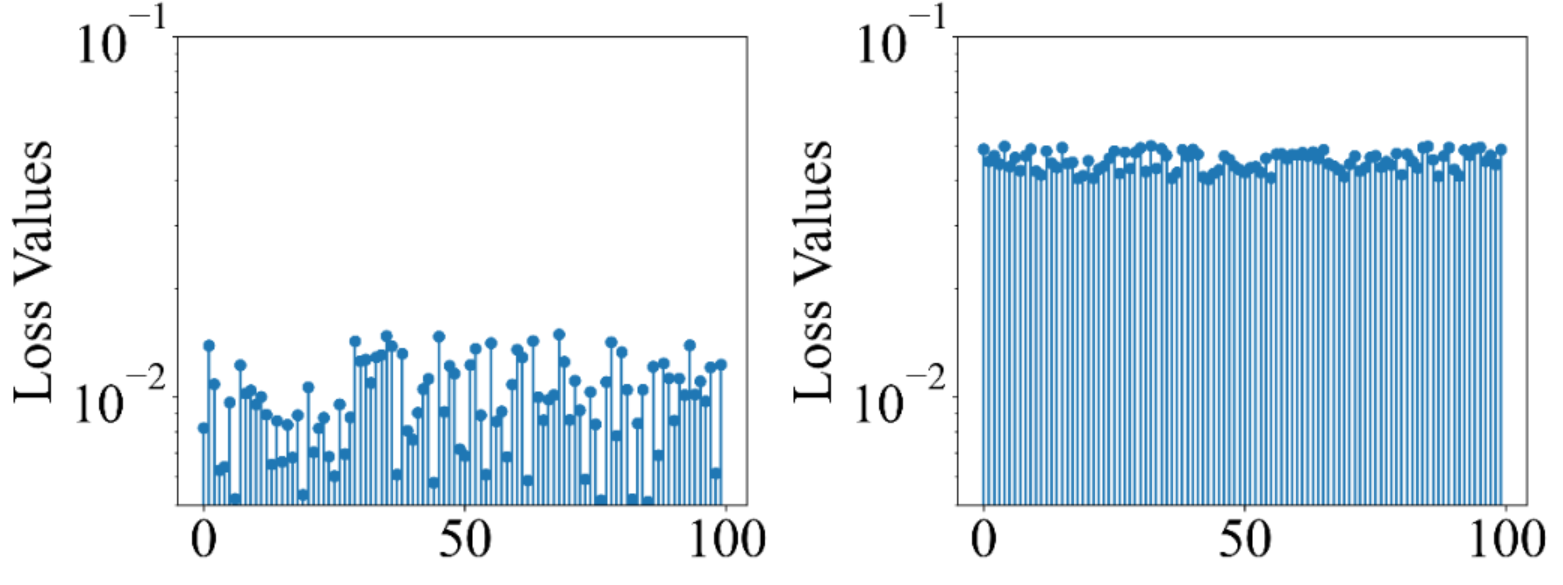}
    \vspace{-1em}
    \caption{HIO to solve FFPR with vs without precise support. We plot the final least-squares losses defined in \cref{eq:LS4PR} over $100$ random starts. $\mb X$ is the groundtruth in \cref{fig:BCDI_sym}. }
    \label{fig:FFPR_HIO}
\end{figure}
the gold-standard hybrid input-output (HIO) algorithm can typically solve the problem when provided with tight support specification\footnote{Specifically, we mean that the tightest rectangular bounding box of the support is provided. }---translation symmetry is killed, but HIO fails when the support is loose---translation symmetry remains; see \cref{fig:FFPR_HIO}. Moreover, our prior works~\cite{TayalEtAl2020Inverse,ManekarEtAl2021Breaking,TayalEtAl2020Unlocking,ManekarEtAl2020Deep} also show the learning difficulties caused by these symmetries when one develops data-driven methods for solving FFPR. See \cref{sec:background} for details of HIO and related algorithms. 

\vspace{0.5em}\noindent\textbf{\textcolor{umn_maroon}{Our focus on practical FFPR methods}} \quad 
We have stressed that symmetries largely determine the difficulty level of PR. However, in previous research, there are often simplifications to FFPR, including \textbf{(1) randomized the model $\mc A$} that only keeps the global phase symmetry, \textbf{(2) evaluation on natural images} that removes the translation symmetry and simplifies the global phase symmetry into sign symmetry~\cite{FannjiangStrohmer2020numerics,LukeEtAl2019Optimization,Marchesini2007Invited}. These simplifications invariably lead to FFPR methods that do not work on practical data. \textbf{The goal of this paper is to develop practical methods for FFPR that involve all three symmetries}. In particular, we propose a novel FFPR method based on double deep image priors (see \cref{sec:method}), and validate its superiority over state-of-the-art (SOTA) on realistic datasets (see \cref{sec:exps}).

\section{FFPR: Formulation and Prior Arts}
\label{sec:background}

\noindent\textbf{\textcolor{umn_maroon}{FFPR model}} \quad 
The object of interest is $\mb X \in \Cp^{m \times n}$, and $\mb Y = \abs{\mc F(\mb X)}^2 \in \R_+^{m'\times n'}$. Here, $\mc F$ is the oversampled 2D Fourier transform: 
\begin{align} \label{eq:oversample}
    \mc F (\mb X) = \mb F_{m'}
    \begin{bmatrix}
    \mb X & \mb 0  \\
    \mb 0 & \mb 0 
    \end{bmatrix}
    \mb F_{n'}^\T,
\end{align}
where $\mb F_{m'}$ and $\mb F_{n'}$ are 
normalized discrete Fourier matrices. We always assume that $m' \ge 2m-1$ and $n' \ge 2n-1$, which is necessary to ensure recoverability.

\vspace{0.5em}\noindent\textbf{\textcolor{umn_maroon}{Prior arts on FFPR}} \quad
Since we focus on practical FFPR, here we only discuss methods that have been tested on FFPR with at least partial success.  
\begin{itemize}[leftmargin=0.15in,itemsep=0.1em] 
    \item \textbf{Classical iterative methods}:  Due to the failure of the LS in \cref{eq:LS4PR}, most (if not all) classical methods tackle the over-parameterized feasibility reformulation: 
    \begin{align} \label{eq:pr_feas_form}
    \text{find}\; \mb Z \in \Cp^{m' \times n'} \; \st \; \abs{\mc F(\mb Z)}^2 = \mb Y, \mc L(\mb Z) = \mb 0, 
    \end{align}
    where $\mc L$ restricts $\mb Z$ to the zero-padding locations defined by the oversampling in \cref{eq:oversample}. \textbf{More refined support information can be naturally incorporated into the support constraint $\mc L(\mb Z) = \mb 0$}. These classical methods are all based on generalized alternating projection for solving \cref{eq:pr_feas_form}, represented by error-reduction (ER), hybrid input-output (HIO)~\cite{fienup1982phase}, reflection average alternating reflectors (RAAR)~\cite{luke2004relaxed}, difference map (DM)~\cite{elser2003phase}, and oversampling smoothness (OSS)~\cite{rodriguez2013oversampling}. They are empirically observed to find good solutions for FFPR, provided that the support specification for $\mb Z$ is tight and hyperparameters are properly tuned. Alternative formulations solved by second-order methods~\cite{marchesini2007phase,Zhuang2020PhaseRV} are less sensitive to hyperparameters. However, all these methods require tight support specification to avoid the translation symmetry---failing so leads to spurious solutions that look like the superposition of translated copies; see \cref{fig:loose_support_bad_sol}.   
    \begin{figure}[!htbp]
        \centering
        \includegraphics[width=0.85\linewidth]{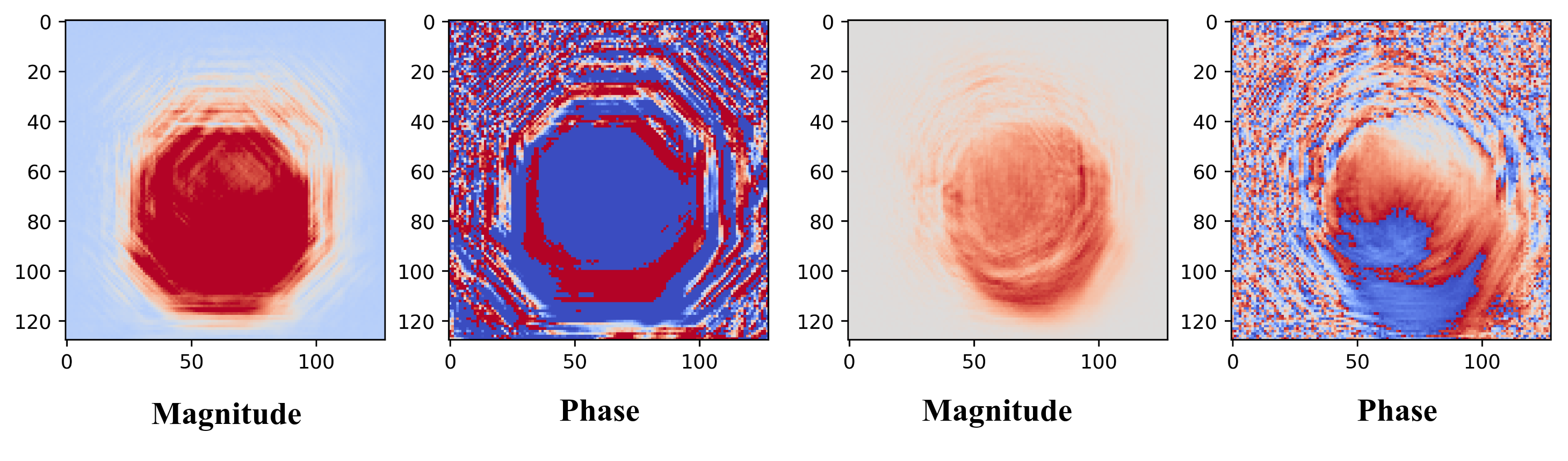}
        \vspace{-1em}
        \caption{Two failure examples when solving FFPR using classical iterative methods without precise support specification and without shrinkwrap. }
        \label{fig:loose_support_bad_sol}
    \end{figure}
    This is addressed by the popular \textbf{shrinkwrap} trick~\cite{MarchesiniEtAl2003X} in practice, which refines the support by smoothing-and-thresholding over iterations. 
    \item \textbf{Data-driven methods}: \textbf{The first line} represents the inverse mapping from $\mb Y$ to $\mb X$ by a deep neural network (DNN) $g_{\mb \theta}$, which is trained either over an extensive training set $\set{(\mb Y_i, \mb X_i)}_i$, or unpaired $\set{\mb Y_i}_i$ and $\set{\mb X_i}_i$ only by observing the cycle consistency constraint: $\abs{\mc F(g_{\mb \theta}(\mb Y))}^2 \approx \mb Y$~\cite{sinha2017lensless,scheinker2020adaptive,chan2021rapid,harder2021deep,wu2021three,yao2022autophasenn,TayalEtAl2020Inverse,ManekarEtAl2021Breaking,TayalEtAl2020Unlocking,ManekarEtAl2020Deep,cherukara2018real, zhang2021phasegan}. But, as discussed in our prior works~\cite{TayalEtAl2020Inverse,ManekarEtAl2021Breaking,TayalEtAl2020Unlocking,ManekarEtAl2020Deep}, symmetries in the problem cause substantial learning difficulties, as any $\mb Y$ maps to a family of equivalent $\mb X$'s. \textbf{The second line} \cite{metzler2018prdeep,icsil2019deep,wang2020deep} is tied to specific iterative methods for solving FFPR and replaces certain components of these methods with trainable DNNs. A common limitation of this line is the reliance on good initialization that is obtained from classical iterative methods. Therefore, this family can be viewed as a final refinement of the results from classical methods and does not address the essential difficulty of solving FFPR. \textbf{Both lines} suffer in generalization when the training data are not sufficiently representative. 
\end{itemize}
Our method overcomes the limitations of both classical and data-driven methods. (1) \textbf{No training set}: it works with a single problem instance each time, with zero extra training data; (2) \textbf{No shrinkwrap}: we can specify the size of $\mb X$ directly as $\lfloor m'/2 \rfloor\times \lfloor n \rfloor$, i.e., the information-theoretic recovery limit, without worrying about the translation symmetry; (3) \textbf{Minimal tuning}: mostly we only need to tune $2$ learning rates as hyperparameters, vs. the $5$ or $6$ hyperparameters used in HIO+ER+Shrinkwrap (HES) for practical coherent diffraction imaging (CDI)~\cite{NewtonEtAl2009Three}.

\section{Our method: FFPR using double DIPs}
\label{sec:method}
\vspace{0.5em}\noindent\textbf{\textcolor{umn_maroon}{Deep image prior (DIP) for visual inverse problems}} \quad 
DIP and variants~\cite{ulyanov2020deep} parameterize visual objects as outputs of DNNs---typically structured convolutional networks to favor spatially smooth structures, i.e., $\mb x = G_{\mb \theta}(\mb z)$, where $\mb z$ is normally a random but fixed seed, and $G_{\mb \theta}$ is a trainable DNN paramaterized by $\mb \theta$. For a visual inverse problem of the form $\mb y \approx f(\mb x)$ where $\mb y$ is the observation and $f$ is the observation model, the classical regularized data-fitting formulation 
\begin{align}
    \min_{\mb x} \; \ell\paren{\mb y, f(\mb x)} + \lambda \Omega(\mb x) 
\end{align}
can now be empowered by DIP and turned into 
\begin{align}
    \min_{\mb \theta} \; \ell\paren{\mb y, f \circ G_{\mb \theta}(\mb z)} + \lambda \Omega \circ G_{\mb \theta}(\mb z). 
\end{align}
This simple idea has recently claimed numerous successes in computer vision and computational imaging; see, e.g., the recent survey~\cite{qayyum2022untrained}, and our recent work addressing practicality issues around DIP~\cite{li2021self,wang2021early,zhuang2022blind}. \textbf{A salient feature of DIP is the strong structured prior it imposes through DNNs, with zero extra data}! Although the theoretical understanding of DIP is still far from complete, current theories attribute its success to two aspects: (1) \textbf{structured priors} imposed by convolutional and upsampling operations, and (2) \textbf{global optimization} due to significant overparameterization and first-order methods~\cite{heckel2019denoising, heckel2020compressive}.  

\begin{figure*}[!htbp]
     \centering
     \includegraphics[width=0.95\linewidth]{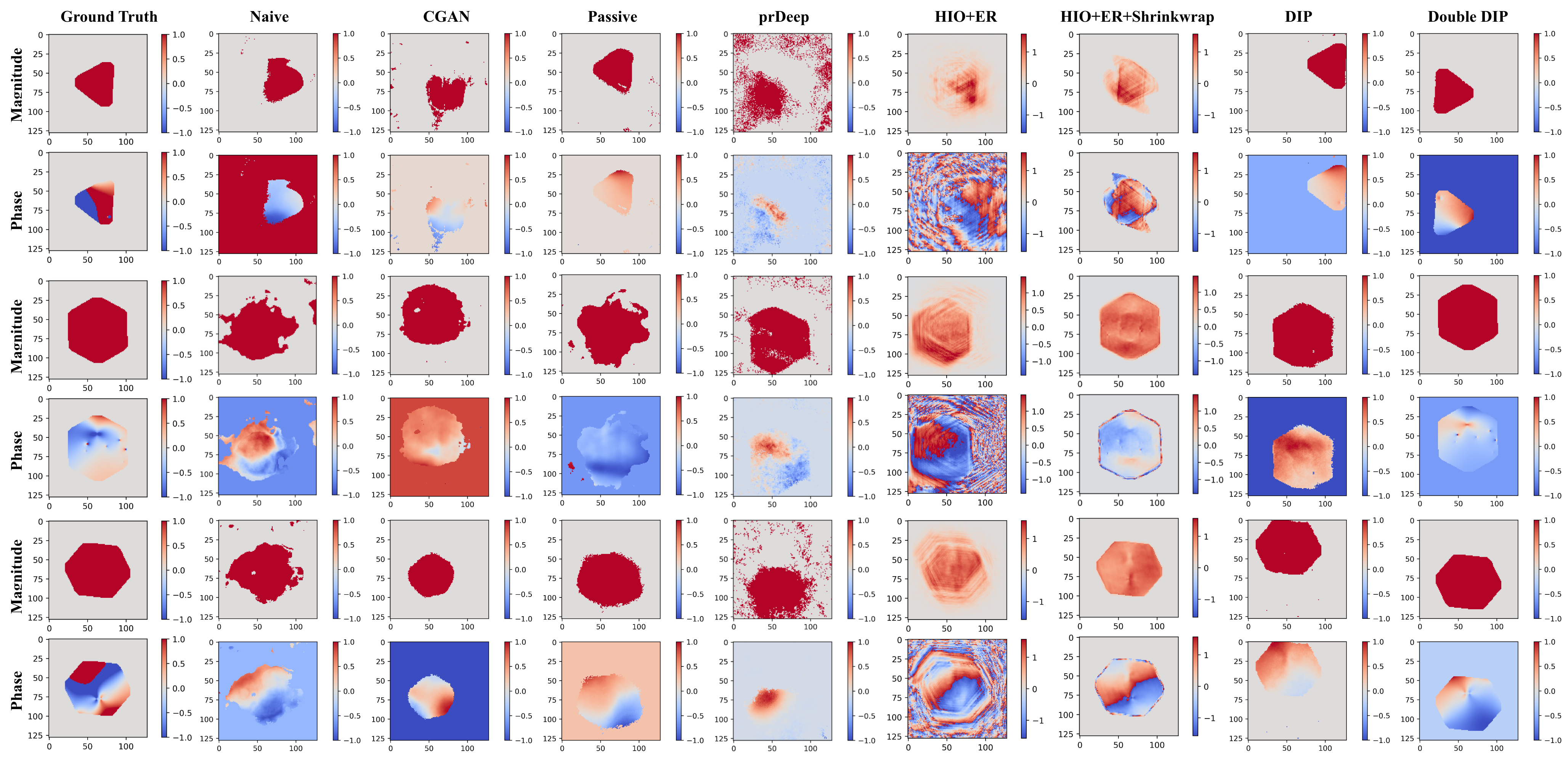}
     \caption{Visual comparison of reconstruction results by different methods on 2D simulated crystal data}
     \label{fig:BCDI_2D_exp}
 \end{figure*}

\begin{figure}[!htbp]
     \centering
     \includegraphics[width=0.7\linewidth]{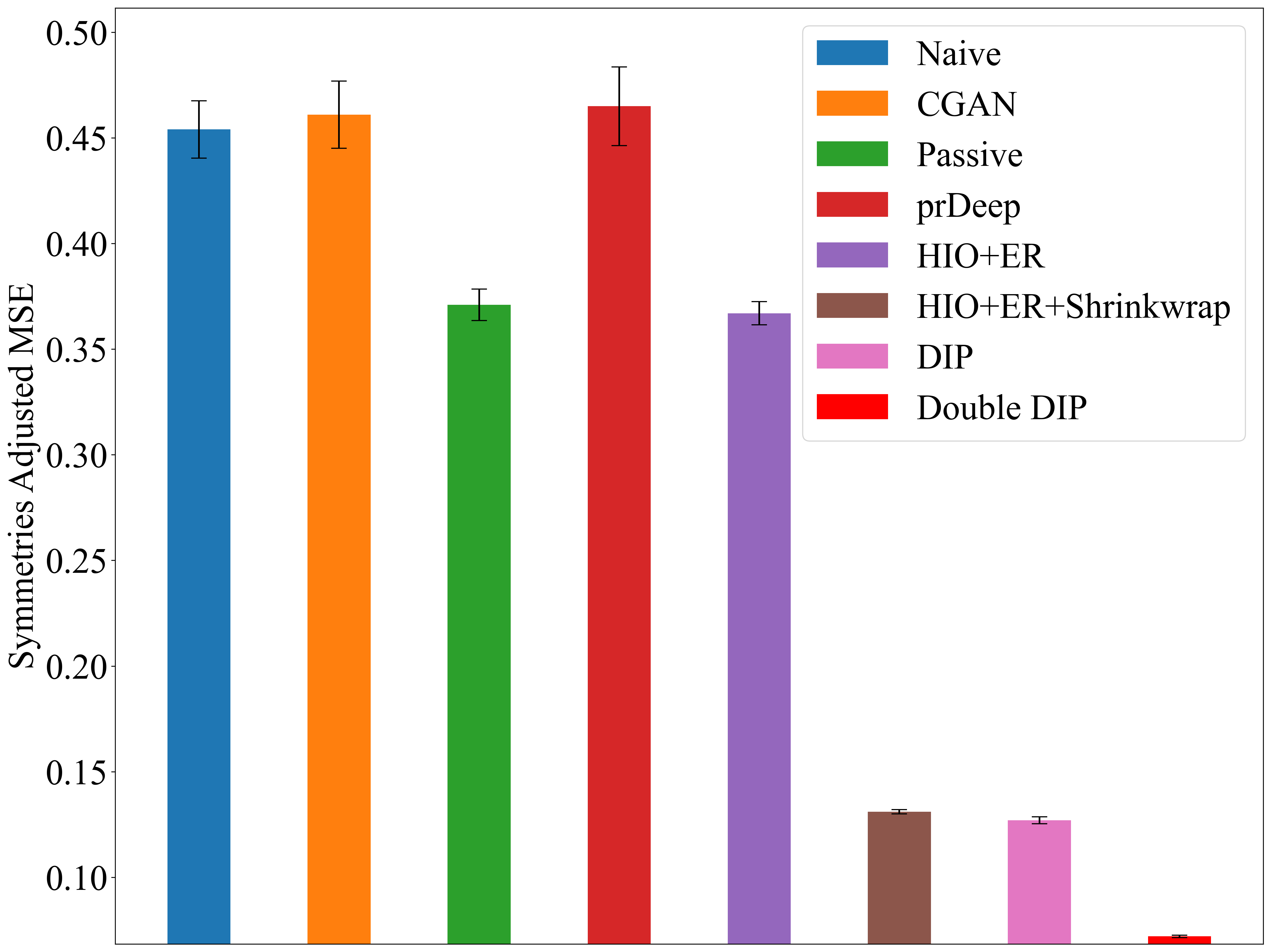}
     \caption{Quantitative comparison of reconstruction results by different methods on 2D simulated crystal data by symmetry-adjusted MSE}
     \label{fig:BCDI_2D_exp_mse}
\end{figure}

\vspace{0.5em}\noindent\textbf{\textcolor{umn_maroon}{Applying DIP to FFPR}} \quad
As shown in \cref{tab:symm_comp}, solving the LS formulation in \cref{eq:LS4PR} using gradient descent always gets trapped in bad local minimizers. It is then tempting to try DIP, as (1) the objects we try to recover in scientific imaging are visual objects and probably can be blessed by the structured priors enforced by DIP, and (2) more importantly, the issue we encounter in solving the LS is exactly about global optimization, which could be eliminated by overparameterization in DIP\footnote{Interestingly, the feasibility reformulation in \cref{eq:pr_feas_form} solved by classical methods can also be understood as performing overparameterization.}. In fact, we have tried the single-DIP version in our preliminary work~\cite{TayalEtAl2021Phase}: 
\begin{align}
      \min_{\mb \theta} \; \|\sqrt{\mb Y} - \abs{\mc F \circ G_{\mb \theta}(\mb z)}\|^2_F,  \quad G_{\mb \theta}(\mb z) \in \Cp^{m \times n}. 
\end{align}
Systematic evaluation in \cref{fig:BCDI_2D_exp,fig:BCDI_2D_exp_mse} shows that it is already competitive compared to the gold-standard HES, although it struggles to reconstruct complicated complex phases.   

\vspace{0.5em}\noindent\textbf{\textcolor{umn_maroon}{Double DIPs boost the performance}} \quad
For FFPR applications such as CDI, $\mb X$ as a complex-valued object can often be naturally split into two parts with disparate complexity levels. For example, in Bragg CDI on crystals, the magnitude part on the support is known to have uniform values, but the phase part can have complex spatial patterns due to strains~\cite{clark2012high,hofmann20173d,yang2022refinements}; in CDI on live cells, the nonnegative real part contains useful information, and the imaginary part acts like small-magnitude noise~\cite{van2015imaging}. In these cases, due to the apparent asymmetry in complexity, it makes sense to parameterize $\mb X$ as two separate DIPs~\cite{GandelsmanEtAl2019Double,zhuang2022blind} instead of one: 
{\small
\begin{align} \label{eq: z2 form}
    \mb X = G^{1}_{\mb \theta_{1}}\paren{\mb z_1} e^{iG^{2}_{\mb \theta_{2}}\paren{\mb z_2}}, \; \text{or}\;  \mb X  =  G^{1}_{\mb \theta_{1}}\paren{\mb z_1} +iG^{2}_{\mb \theta_{2}}\paren{\mb z_2}. 
\end{align}} 
This can be justified as balancing the learning paces: with a single DIP, ``simple" part is learned much faster than the ``complex'' part; with double DIPs, we can balance the learning paces by making the learning rate for the ``simple" part relatively small compared to that for the ``complex" part. We observe a substantial performance boost in \cref{fig:BCDI_2D_exp,fig:BCDI_2D_exp_mse} due to the double-DIP parametrization.

\begin{figure}[!htbp]
     \centering
     \includegraphics[width=0.7\linewidth]{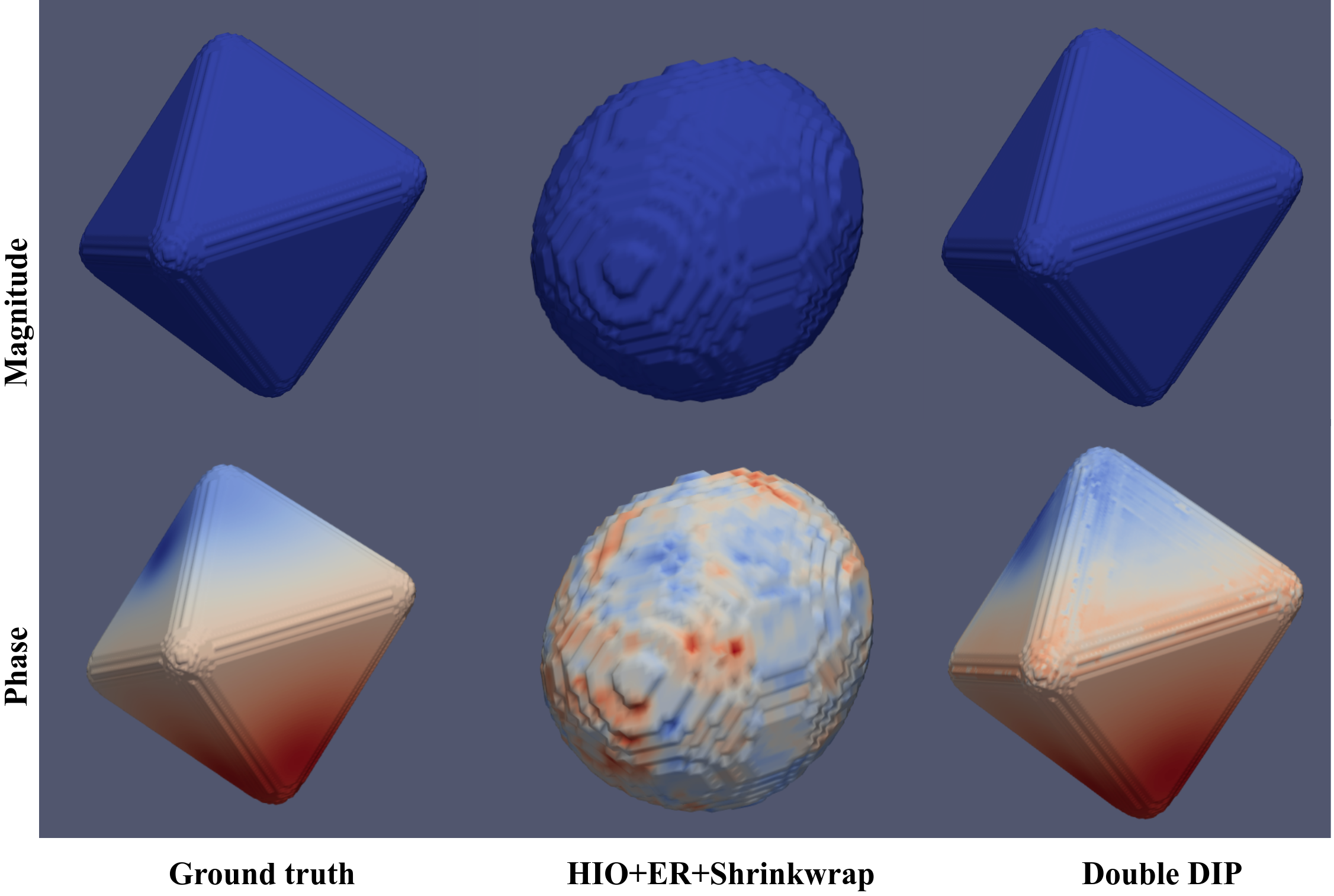}
     \caption{Visual comparison of reconstruction results by HIO+ER+Shrinkwrap and our method on a 3D simulated crystal instance}
     \label{fig:BCDI_3D_exp}
\end{figure}
\section{Experiments Results}
\label{sec:exps}
\vspace{0.5em}\noindent\textbf{\textcolor{umn_maroon}{Evaluation on 2D simulated Bragg CDI crystal data}} \quad 
We first compare our Double-DIP method with multiple SOTA methods for FFPR, including Naive~\cite{cherukara2018real}, CGAN~\cite{uelwer2021phase}, Passive~\cite{chan2021rapid}, prDeep~\cite{metzler2018prdeep}, HIO+ER, HIO+ER+Shrinkwrap (HES), and (single-)DIP on simulated 2D data for Bragg CDI on crystals. The final form of our learning objective for this task is: 
\begin{multline}
   \min_{\mb \theta_1, \mb \theta_2} \; \|\sqrt{\mb Y} - |\mc F \circ  G^{1}_{\mb \theta_{1}}\paren{\mb z_1} e^{iG^{2}_{\mb \theta_{2}}\paren{\mb z_2}}|\|^2_F,  \\
   G^{1}_{\mb \theta_{1}}\paren{\mb z_1} \in \Cp^{m \times n}, \; G^{2}_{\mb \theta_{2}}\paren{\mb z_2} \in \Cp^{m \times n}. 
\end{multline}
To ensure that the evaluation data reflect real-world complexity, we simulate 2D complex-valued crystal data in Bragg CDI applications~\cite{RobinsonEtAl2001Reconstruction}. The dataset is generated by first creating 2D convex and nonconvex shapes based on random scattering points in a $110\times 110$ grid on a $128 \times 128$ background. The complex magnitudes are uniformly $1$, and the complex phases are determined by projecting the simulated 2D displacement fields (due to crystal defects) onto the corresponding momentum transfer vectors. To maximize the diversity, the dataset contains diverse shapes and different numbers and densities of crystal defects that directly determine the complexity of the phases. \textbf{Although our double-DIP method is a  single-instance method that requires no training data, the dataset is large enough to support data-driven methods, such as Passive and prDeep}. For methods that require a training set, we provide $8000$ samples. All methods are tested on $50$ samples. \\
\\
From both visual (\cref{fig:BCDI_2D_exp}) and quantitative (\cref{fig:BCDI_2D_exp_mse}) results, it is evident that: \textbf{(1)} all data-driven methods, including Naive, CGAN, Passive, prDeep, perform poorly. We believe that this is due to either the learning difficulty caused by the three symmetries~\cite{TayalEtAl2020Inverse,ManekarEtAl2021Breaking,TayalEtAl2020Unlocking,ManekarEtAl2020Deep} or the bad initialization given by HIO (i.e., for prDeep); \textbf{(2)} HES, DIP, and our double-DIP are the top three methods. HES deals with translation symmetry by explicitly iterative refining the support, whereas DIP and ours do not need tight support estimation at all, substantially reducing parameter tuning; \textbf{(3)} Our method wins HES and DIP by a large margin. Although the latter two perform reasonably well in magnitude estimation, their phase estimations are typically off for complicated instances. 

\vspace{0.5em}\noindent\textbf{\textcolor{umn_maroon}{Evaluation on 3D simulated Bragg CDI crystal data}} \quad 
We will not continue considering data-driven methods, due to their clear performance deficiency on 2D data and the considerable cost to obtain sufficiently representative training sets for 3D. We only compare HES, which is the gold-standard used in Bragg CDI practice, with our double-DIP method. Since both methods can work with single instances and need no training data, here we quickly compare their performance qualitatively on a single 3D simulated crystal instance (the simulation process is similar to the 2D case), as shown in \cref{fig:BCDI_3D_exp}. It is obvious that even with Shrinkwrap iteratively refining support, HES still struggles to get the support right. By contrast, our double-DIP method obtains sharp support recovery and good phase estimation.

{\small  
\bibliographystyle{IEEEbib}
\bibliography{icassp}

\begin{thebibliography}{10}

\bibitem{Goodman2017Introduction}
Joseph~W. Goodman,
\newblock {\em Introduction to Fourier Optics},
\newblock Freeman \& Company, W. H., 4th edition, 2017.

\bibitem{miao1999extending}
Jianwei Miao, Pambos Charalambous, Janos Kirz, and David Sayre,
\newblock ``Extending the methodology of x-ray crystallography to allow imaging
  of micrometre-sized non-crystalline specimens,''
\newblock {\em Nature}, vol. 400, no. 6742, pp. 342--344, 1999.

\bibitem{RobinsonEtAl2001Reconstruction}
I.~K. Robinson, I.~A. Vartanyants, G.~J. Williams, M.~A. Pfeifer, and J.~A.
  Pitney,
\newblock ``Reconstruction of the shapes of gold nanocrystals using coherent
  x-ray diffraction,''
\newblock {\em Physical Review Letters}, vol. 87, no. 19, pp. 195505, oct 2001.

\bibitem{LukeEtAl2002Optical}
D.~Russell Luke, James~V. Burke, and Richard~G. Lyon,
\newblock ``Optical wavefront reconstruction: Theory and numerical methods,''
\newblock {\em {SIAM} Review}, vol. 44, no. 2, pp. 169--224, jan 2002.

\bibitem{Jaming1999Phase}
Philippe Jaming,
\newblock ``Phase retrieval techniques for radar ambiguity problems,''
\newblock {\em The Journal of Fourier Analysis and Applications}, vol. 5, no.
  4, pp. 309--329, jul 1999.

\bibitem{ShechtmanEtAl2015Phase}
Yoav Shechtman, Yonina~C. Eldar, Oren Cohen, Henry~Nicholas Chapman, Jianwei
  Miao, and Mordechai Segev,
\newblock ``Phase retrieval with application to optical imaging: A contemporary
  overview,''
\newblock {\em {IEEE} Signal Processing Magazine}, vol. 32, no. 3, pp. 87--109,
  may 2015.

\bibitem{candes2015phase}
Emmanuel~J Candes, Yonina~C Eldar, Thomas Strohmer, and Vladislav Voroninski,
\newblock ``Phase retrieval via matrix completion,''
\newblock {\em SIAM review}, vol. 57, no. 2, pp. 225--251, 2015.

\bibitem{FannjiangStrohmer2020numerics}
Albert Fannjiang and Thomas Strohmer,
\newblock ``The numerics of phase retrieval,''
\newblock {\em Acta Numerica}, vol. 29, pp. 125--228, may 2020.

\bibitem{WangEtAl2020Phase}
Fei Wang, Yaoming Bian, Haichao Wang, Meng Lyu, Giancarlo Pedrini, Wolfgang
  Osten, George Barbastathis, and Guohai Situ,
\newblock ``Phase imaging with an untrained neural network,''
\newblock {\em Light: Science \& Applications}, vol. 9, no. 1, may 2020.

\bibitem{ZhangEtAl2021PhaseGAN}
Yuhe Zhang, Mike~Andreas Noack, Patrik Vagovic, Kamel Fezzaa, Francisco
  Garcia-Moreno, Tobias Ritschel, and Pablo Villanueva-Perez,
\newblock ``{PhaseGAN}: a deep-learning phase-retrieval approach for unpaired
  datasets,''
\newblock {\em Optics Express}, vol. 29, no. 13, pp. 19593, jun 2021.

\bibitem{ShimobabaIto2019Computera}
Tomoyoshi Shimobaba and Tomoyoshi Ito,
\newblock {\em Computer Holography Acceleration Algorithms and Hardware
  Implementations},
\newblock Taylor \& Francis Group, 2019.

\bibitem{Fienup1993Phase}
J.~R. Fienup,
\newblock ``Phase-retrieval algorithms for a complicated optical system,''
\newblock {\em Applied Optics}, vol. 32, no. 10, pp. 1737, apr 1993.

\bibitem{BarmherzigSun2022Towards}
David~A. Barmherzig and Ju~Sun,
\newblock ``Towards practical holographic coherent diffraction imaging via
  maximum likelihood estimation,''
\newblock {\em Optics Express}, vol. 30, no. 5, pp. 6886, feb 2022.

\bibitem{ThibaultEtAl2008High}
Pierre Thibault, Martin Dierolf, Andreas Menzel, Oliver Bunk, Christian David,
  and Franz Pfeiffer,
\newblock ``High-resolution scanning x-ray diffraction microscopy,''
\newblock {\em Science}, vol. 321, no. 5887, pp. 379--382, jul 2008.

\bibitem{MarchesiniEtAl2013Augmented}
Stefano Marchesini, Andre Schirotzek, Chao Yang, Hau tieng Wu, and Filipe Maia,
\newblock ``Augmented projections for ptychographic imaging,''
\newblock {\em Inverse Problems}, vol. 29, no. 11, pp. 115009, oct 2013.

\bibitem{BendoryEtAl2017Fourier}
Tamir Bendory, Robert Beinert, and Yonina~C. Eldar,
\newblock ``Fourier phase retrieval: Uniqueness and algorithms,''
\newblock in {\em Compressed Sensing and its Applications}, pp. 55--91.
  Springer International Publishing, 2017.

\bibitem{yeh2015experimental}
Li-Hao Yeh, Jonathan Dong, Jingshan Zhong, Lei Tian, Michael Chen, Gongguo
  Tang, Mahdi Soltanolkotabi, and Laura Waller,
\newblock ``Experimental robustness of fourier ptychography phase retrieval
  algorithms,''
\newblock {\em Optics express}, vol. 23, no. 26, pp. 33214--33240, 2015.

\bibitem{TayalEtAl2020Inverse}
Kshitij Tayal, Chieh-Hsin Lai, Vipin Kumar, and Ju~Sun,
\newblock ``Inverse problems, deep learning, and symmetry breaking,''
\newblock {\em arXiv:2003.09077}, Mar. 2020.

\bibitem{ManekarEtAl2021Breaking}
Raunak Manekar, Kshitij Tayal, Zhong Zhuang, Chieh-Hsin Lai, Vipin Kumar, and
  Ju~Sun,
\newblock ``Breaking symmetries in data-driven phase retrieval,''
\newblock in {\em {OSA} Imaging and Applied Optics Congress 2021 (3D, {COSI},
  {DH}, {ISA}, {pcAOP})}. 2021, Optica Publishing Group.

\bibitem{TayalEtAl2020Unlocking}
Kshitij Tayal, Chieh-Hsin Lai, Raunak Manekar, Zhong Zhuang, Vipin Kumar, and
  Ju~Sun,
\newblock ``Unlocking inverse problems using deep learning: Breaking symmetries
  in phase retrieval,''
\newblock in {\em NeurIPS 2020 Workshop on Deep Learning and Inverse Problems},
  2020.

\bibitem{ManekarEtAl2020Deep}
Raunak Manekar, Zhong Zhuang, Kshitij Tayal, Vipin Kumar, and Ju~Sun,
\newblock ``Deep learning initialized phase retrieval,''
\newblock in {\em NeurIPS 2020 Workshop on Deep Learning and Inverse Problems},
  2020.

\bibitem{LukeEtAl2019Optimization}
D.~Russell Luke, Shoham Sabach, and Marc Teboulle,
\newblock ``Optimization on spheres: Models and proximal algorithms with
  computational performance comparisons,''
\newblock {\em {SIAM} Journal on Mathematics of Data Science}, vol. 1, no. 3,
  pp. 408--445, jan 2019.

\bibitem{Marchesini2007Invited}
S.~Marchesini,
\newblock ``Invited article: A unified evaluation of iterative projection
  algorithms for phase retrieval,''
\newblock {\em Review of Scientific Instruments}, vol. 78, no. 1, pp. 011301,
  jan 2007.

\bibitem{fienup1982phase}
James~R Fienup,
\newblock ``Phase retrieval algorithms: a comparison,''
\newblock {\em Applied optics}, vol. 21, no. 15, pp. 2758--2769, 1982.

\bibitem{luke2004relaxed}
D~Russell Luke,
\newblock ``Relaxed averaged alternating reflections for diffraction imaging,''
\newblock {\em Inverse problems}, vol. 21, no. 1, pp. 37, 2004.

\bibitem{elser2003phase}
Veit Elser,
\newblock ``Phase retrieval by iterated projections,''
\newblock {\em JOSA A}, vol. 20, no. 1, pp. 40--55, 2003.

\bibitem{rodriguez2013oversampling}
Jose~A Rodriguez, Rui Xu, C-C Chen, Yunfei Zou, and Jianwei Miao,
\newblock ``Oversampling smoothness: an effective algorithm for phase retrieval
  of noisy diffraction intensities,''
\newblock {\em Journal of applied crystallography}, vol. 46, no. 2, pp.
  312--318, 2013.

\bibitem{marchesini2007phase}
Stefano Marchesini,
\newblock ``Phase retrieval and saddle-point optimization,''
\newblock {\em JOSA A}, vol. 24, no. 10, pp. 3289--3296, 2007.

\bibitem{Zhuang2020PhaseRV}
Zhong Zhuang, Gang Wang, Yash Travadi, and Ju~Sun,
\newblock ``Phase retrieval via second-order nonsmooth optimization,''
\newblock in {\em ICML Workshop on Beyond First-Order Methods for Machine
  Learning}, 2020.

\bibitem{MarchesiniEtAl2003X}
S.~Marchesini, H.~He, H.~N. Chapman, S.~P. Hau-Riege, A.~Noy, M.~R. Howells,
  U.~Weierstall, and J.~C.~H. Spence,
\newblock ``X-ray image reconstruction from a diffraction pattern alone,''
\newblock {\em Physical Review B}, vol. 68, no. 14, pp. 140101, oct 2003.

\bibitem{sinha2017lensless}
Ayan Sinha, Justin Lee, Shuai Li, and George Barbastathis,
\newblock ``Lensless computational imaging through deep learning,''
\newblock {\em Optica}, vol. 4, no. 9, pp. 1117--1125, 2017.

\bibitem{scheinker2020adaptive}
Alexander Scheinker and Reeju Pokharel,
\newblock ``Adaptive 3d convolutional neural network-based reconstruction
  method for 3d coherent diffraction imaging,''
\newblock {\em Journal of Applied Physics}, vol. 128, no. 18, pp. 184901, 2020.

\bibitem{chan2021rapid}
Henry Chan, Youssef~SG Nashed, Saugat Kandel, Stephan~O Hruszkewycz,
  Subramanian~KRS Sankaranarayanan, Ross~J Harder, and Mathew~J Cherukara,
\newblock ``Rapid 3d nanoscale coherent imaging via physics-aware deep
  learning,''
\newblock {\em Applied Physics Reviews}, vol. 8, no. 2, pp. 021407, 2021.

\bibitem{harder2021deep}
Ross Harder,
\newblock ``Deep neural networks in real-time coherent diffraction imaging,''
\newblock {\em IUCrJ}, vol. 8, no. Pt 1, pp. 1, 2021.

\bibitem{wu2021three}
Longlong Wu, Shinjae Yoo, Ana~F Suzana, Tadesse~A Assefa, Jiecheng Diao, Ross~J
  Harder, Wonsuk Cha, and Ian~K Robinson,
\newblock ``Three-dimensional coherent x-ray diffraction imaging via deep
  convolutional neural networks,''
\newblock {\em npj Computational Materials}, vol. 7, no. 1, pp. 1--8, 2021.

\bibitem{yao2022autophasenn}
Yudong Yao, Henry Chan, Subramanian Sankaranarayanan, Prasanna Balaprakash,
  Ross~J Harder, and Mathew~J Cherukara,
\newblock ``Autophasenn: unsupervised physics-aware deep learning of 3d
  nanoscale bragg coherent diffraction imaging,''
\newblock {\em npj Computational Materials}, vol. 8, no. 1, pp. 1--8, 2022.

\bibitem{cherukara2018real}
Mathew~J Cherukara, Youssef~SG Nashed, and Ross~J Harder,
\newblock ``Real-time coherent diffraction inversion using deep generative
  networks,''
\newblock {\em Scientific reports}, vol. 8, no. 1, pp. 1--8, 2018.

\bibitem{zhang2021phasegan}
Yuhe Zhang, Mike~Andreas Noack, Patrik Vagovic, Kamel Fezzaa, Francisco
  Garcia-Moreno, Tobias Ritschel, and Pablo Villanueva-Perez,
\newblock ``Phasegan: a deep-learning phase-retrieval approach for unpaired
  datasets,''
\newblock {\em Optics express}, vol. 29, no. 13, pp. 19593--19604, 2021.

\bibitem{metzler2018prdeep}
Christopher Metzler, Phillip Schniter, Ashok Veeraraghavan, et~al.,
\newblock ``prdeep: Robust phase retrieval with a flexible deep network,''
\newblock in {\em International Conference on Machine Learning}. PMLR, 2018,
  pp. 3501--3510.

\bibitem{icsil2019deep}
{\c{C}}a{\u{g}}atay I{\c{s}}{\i}l, Figen~S Oktem, and Aykut Ko{\c{c}},
\newblock ``Deep iterative reconstruction for phase retrieval,''
\newblock {\em Applied optics}, vol. 58, no. 20, pp. 5422--5431, 2019.

\bibitem{wang2020deep}
Yaotian Wang, Xiaohang Sun, and Jason Fleischer,
\newblock ``When deep denoising meets iterative phase retrieval,''
\newblock in {\em International Conference on Machine Learning}. PMLR, 2020,
  pp. 10007--10017.

\bibitem{NewtonEtAl2009Three}
Marcus~C. Newton, Steven~J. Leake, Ross Harder, and Ian~K. Robinson,
\newblock ``Three-dimensional imaging of strain in a single {ZnO} nanorod,''
\newblock {\em Nature Materials}, vol. 9, no. 2, pp. 120--124, dec 2009.

\bibitem{ulyanov2020deep}
Dmitry Ulyanov, Andrea Vedaldi, and Victor Lempitsky,
\newblock ``Deep image prior,''
\newblock {\em International Journal of Computer Vision}, vol. 128, no. 7, pp.
  1867--1888, 2020.

\bibitem{qayyum2022untrained}
Adnan Qayyum, Inaam Ilahi, Fahad Shamshad, Farid Boussaid, Mohammed Bennamoun,
  and Junaid Qadir,
\newblock ``Untrained neural network priors for inverse imaging problems: A
  survey,''
\newblock {\em IEEE Transactions on Pattern Analysis and Machine Intelligence},
  2022.

\bibitem{li2021self}
Taihui Li, Zhong Zhuang, Hengyue Liang, Le~Peng, Hengkang Wang, and Ju~Sun,
\newblock ``Self-validation: Early stopping for single-instance deep generative
  priors,''
\newblock {\em arXiv preprint arXiv:2110.12271}, 2021.

\bibitem{wang2021early}
Hengkang Wang, Taihui Li, Zhong Zhuang, Tiancong Chen, Hengyue Liang, and
  Ju~Sun,
\newblock ``Early stopping for deep image prior,''
\newblock {\em arXiv preprint arXiv:2112.06074}, 2021.

\bibitem{zhuang2022blind}
Zhong Zhuang, Taihui Li, Hengkang Wang, and Ju~Sun,
\newblock ``Blind image deblurring with unknown kernel size and substantial
  noise,''
\newblock {\em arXiv preprint arXiv:2208.09483}, 2022.

\bibitem{heckel2019denoising}
Reinhard Heckel and Mahdi Soltanolkotabi,
\newblock ``Denoising and regularization via exploiting the structural bias of
  convolutional generators,''
\newblock {\em arXiv preprint arXiv:1910.14634}, 2019.

\bibitem{heckel2020compressive}
Reinhard Heckel and Mahdi Soltanolkotabi,
\newblock ``Compressive sensing with un-trained neural networks: Gradient
  descent finds a smooth approximation,''
\newblock in {\em International Conference on Machine Learning}. PMLR, 2020,
  pp. 4149--4158.

\bibitem{TayalEtAl2021Phase}
Kshitij Tayal, Raunak Manekar, Zhong Zhuang, David Yang, Vipin Kumar, Felix
  Hofmann, and Ju~Sun,
\newblock ``Phase retrieval using single-instance deep generative prior,''
\newblock in {\em {OSA} Optical Sensors and Sensing Congress 2021 ({AIS},
  {FTS}, {HISE}, {SENSORS}, {ES})}. 2021, Optica Publishing Group.

\bibitem{clark2012high}
JN~Clark, X~Huang, R~Harder, and IK~Robinson,
\newblock ``High-resolution three-dimensional partially coherent diffraction
  imaging,''
\newblock {\em Nature communications}, vol. 3, no. 1, pp. 1--6, 2012.

\bibitem{hofmann20173d}
Felix Hofmann, Edmund Tarleton, Ross~J Harder, Nicholas~W Phillips, Pui-Wai Ma,
  Jesse~N Clark, Ian~K Robinson, Brian Abbey, Wenjun Liu, and Christian~E Beck,
\newblock ``3d lattice distortions and defect structures in ion-implanted
  nano-crystals,''
\newblock {\em Scientific reports}, vol. 7, no. 1, pp. 1--10, 2017.

\bibitem{yang2022refinements}
David Yang, Mark~T Lapington, Guanze He, Kay Song, Minyi Zhang, Clara Barker,
  Ross~J Harder, Wonsuk Cha, Wenjun Liu, Nicholas~W Phillips, et~al.,
\newblock ``Refinements for bragg coherent x-ray diffraction imaging: Electron
  backscatter diffraction alignment and strain field computation,''
\newblock {\em arXiv preprint arXiv:2203.17015}, 2022.

\bibitem{van2015imaging}
Gijs Van Der~Schot, Martin Svenda, Filipe~RNC Maia, Max Hantke, Daniel~P
  DePonte, M~Marvin Seibert, Andrew Aquila, Joachim Schulz, Richard Kirian,
  Mengning Liang, et~al.,
\newblock ``Imaging single cells in a beam of live cyanobacteria with an x-ray
  laser,''
\newblock {\em Nature communications}, vol. 6, no. 1, pp. 1--9, 2015.

\bibitem{GandelsmanEtAl2019Double}
Yosef Gandelsman, Assaf Shocher, and Michal Irani,
\newblock ``{\textquotedblleft}double-{DIP}{\textquotedblright}: Unsupervised
  image decomposition via coupled deep-image-priors,''
\newblock in {\em 2019 {IEEE}/{CVF} Conference on Computer Vision and Pattern
  Recognition ({CVPR})}. jun 2019, {IEEE}.

\bibitem{uelwer2021phase}
Tobias Uelwer, Alexander Oberstra{\ss}, and Stefan Harmeling,
\newblock ``Phase retrieval using conditional generative adversarial
  networks,''
\newblock in {\em 2020 25th International Conference on Pattern Recognition
  (ICPR)}. IEEE, 2021, pp. 731--738.

\bibitem{ShimobabaIto2019Computer}
Tomoyoshi Shimobaba and Tomoyoshi Ito,
\newblock {\em Computer Holography Acceleration Algorithms and Hardware
  Implementations},
\newblock Taylor \& Francis Group, 2019.

\end{thebibliography}
} 

\clearpage 

\onecolumn
\appendices
\section{Fresnel and Fraunhofer approximations to the diffraction formula}
\label{sec:diff_approx_explained}
\begin{figure}[!htbp]
    \centering 
    \includegraphics[width=0.5\textwidth]{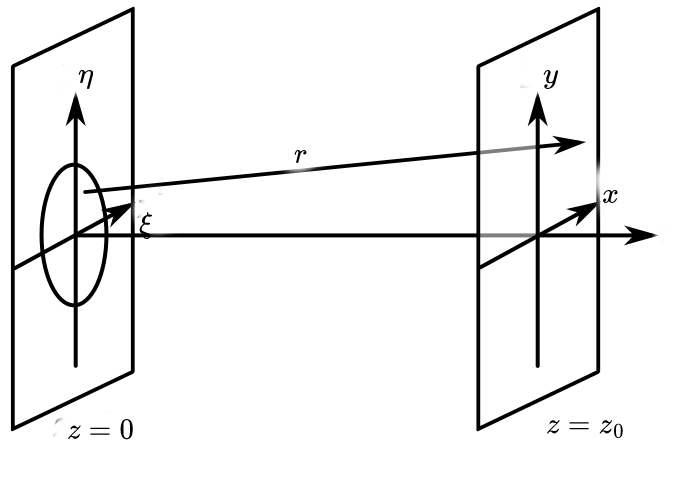}
    \caption{A schematic diffraction imaging system with parallel aperture and imaging planes (plot adapted from \protect\url{https://commons.wikimedia.org/wiki/File:Diffraction_geometry.svg} under the Creative Commons Attribution-Share Alike 3.0 Unported license.)}
    \label{fig:diffraction_pic}
\end{figure}
In this section, we clarify the difference between the near-field and far-field models for phase retrieval, following~\cite{Goodman2017Introduction,ShimobabaIto2019Computer}. Consider the propagation of a monochromatic wave (with wavelength $\lambda$) from an aperture plane $z = 0$ to a parallel imaging plane $z = z_0$ with $z_0 \gg \lambda$; see \cref{fig:diffraction_pic}. Let $U(x, y, z)$ denote the wave field. The celebrated Rayleigh-Sommerfeld diffraction formula dictates that 
\begin{align} \label{eq:RS_spatial}
    \boxed{U(x, y, z_0) = \frac{1}{2\pi}\iint U(\xi, \eta, 0) \paren{\frac{1}{r\paren{\xi, \eta}} - ik} \frac{z_0}{r^2\paren{\xi, \eta}} e^{ikr\paren{\xi, \eta}}\; d \xi d\eta},   
\end{align}
where $r\paren{\xi, \eta} \doteq \sqrt{z_0^2 + (x-\xi)^2 + (y -\eta)^2}$, $k \doteq \frac{2\pi}{\lambda}$ is the wavenumber, and the effective domain of the double integral is the aperture $\Omega$. Moreover, write the domain of the image plane as $\Sigma \subset \R^2$. \cref{eq:RS_spatial} has an equivalent form: 
\begin{align} \label{eq:RS_fourier}
   \boxed{ U(x, y, z_0) = \iint \wh{U}\paren{f_X, f_Y, 0} e^{i k z_0\sqrt{1 - \paren{\lambda f_X}^2 - \paren{\lambda f_Y}^2}} e^{i 2\pi \paren{f_X x + f_Y y}}\; df_X df_Y},   
\end{align}  
where $\wh{U}\paren{f_X, f_Y, 0} = \iint U(x, y, 0) e^{-i 2\pi \paren{f_X x + f_Y y}}\; df_X df_Y$ is the 2D Fourier transform of the planar field $U(x, y, 0)$. The equivalence is due to the convolution theorem: write $r^\circ\paren{x, y} \doteq \sqrt{z_0^2 + x^2 + y^2}$ and 
$h(x, y, z_0) \doteq \frac{1}{2\pi} (\frac{1}{r^\circ(x, y)} - ik) \frac{z_0}{[r^\circ(x, y)]^2} e^{ikr^\circ(x, y)}$. Then \cref{eq:RS_spatial} can be written as $U(x, y, z_0) = U(x, y, 0) \ast h(x,y, z_0)$. The equivalence is clear once we recognize that $\wh{r^\circ}\paren{f_X, f_Y, z_0} \doteq \iint h(x,y, z_0) e^{-i 2\pi \paren{f_X x + f_Y y}}\; df_X df_Y = e^{i k z_0 \sqrt{1- (\lambda f_X)^2 - \paren{\lambda  f_Y}^2} }$. 

For \cref{eq:RS_spatial}, $\frac{1}{r\paren{\xi, \eta}} - ik = \frac{1}{r\paren{\xi, \eta}} - \frac{i 2\pi}{\lambda} \approx \frac{2\pi}{i\lambda}$ as $r\paren{\xi, \eta} \gg \lambda$ in practice. Hence we have the simplified form called the \textbf{Huygens-Fresnel principle}: 
\begin{align} \label{eq:HF_spatial}
    \boxed{U(x, y, z_0) = \frac{z_0}{i\lambda}\iint U(\xi, \eta, 0) \frac{1}{r^2\paren{\xi, \eta}} e^{ikr\paren{\xi, \eta}}\; d \xi d\eta}.   
\end{align}
We can derive two approximations to \cref{eq:HF_spatial}, i.e., the Fresnel (i.e., near-field) and Fraunhofer (i.e., far-field) approximations. Noting that $\sqrt{1+\eps} = 1 + \eps/2 - \eps^2/8 + O(\eps^3)$ for $\abs{\eps} \ll 1$, we can approximate 
\begin{align} 
    r\paren{\xi, \eta} \doteq \sqrt{z_0^2 + (x-\xi)^2 + (y -\eta)^2} = z_0 \sqrt{1+\paren{\frac{x - \xi}{z_0}}^2 + \paren{\frac{y - \eta}{z_0}}^2}   
\end{align}
by its low-order Taylor expansions, provided that $\paren{x - \xi}^2 + \paren{y - \eta}^2 \ll z_0^2$ for all $\paren{\xi, \eta} \in \Omega$. First, we have $r^2\paren{\xi, \eta} \approx z_0^2$ using only the $0$-th order expansion. For the exponential term, since $k$ is normally large, we use the $1$-st order expansion: 
\begin{align}  \label{eq:near-second-approx}
    \exp\paren{ikr\paren{\xi, \eta}} & \approx \exp\brac{ik\paren{z_0 + \frac{(x-\xi)^2}{2z_0} + \frac{(y -\eta)^2}{2z_0}}} = e^{ikz_0} \exp\brac{\frac{ik}{2z_0}\paren{(x- \xi)^2 + (y - \eta)^2}}, 
\end{align} 
which is acceptable when 
\begin{align} 
\frac{z_0 k}{8} \paren{\paren{\frac{x-\xi}{z_0}}^2 + \paren{\frac{y-\eta}{z_0}}^2}^2 \ll 1 \Longleftrightarrow z_0^3 \gg \frac{k}{8} \brac{\paren{x-\xi}^2 + \paren{y-\eta}^2}^2    \quad \forall\, (\xi, \eta) \in \Omega, (x, y) \in \Sigma. 
\end{align}
So we arrive at the famous Fresnel approximation 
\begin{empheq}[innerbox=\fbox,right= \; .\quad (\textbf{Fresnel approximation---forward})]{align*}  \label{eq:fresnel_approx}
    U(x, y, z_0) & \approx \frac{e^{ikz_0}}{i \lambda z_0 } \iint U(\xi, \eta, 0)   \exp\brac{\frac{ik}{2z_0}\paren{(x- \xi)^2 + (y - \eta)^2}}\; d \xi d\eta \\
    & = \frac{e^{ikz_0}}{i \lambda z_0} e^{\frac{ik}{2z_0} \paren{x^2 + y^2}}\iint \brac{U(\xi, \eta, 0) e^{\frac{ik}{2z_0} \paren{\xi^2 + \eta^2}}} e^{-\frac{ik}{z_0}(x\xi + y \eta)}  \; d \xi d\eta
\end{empheq} 
If moreover $\frac{ik}{2z_0} \paren{\xi^2 + \eta^2} \ll 1 \Longleftrightarrow z_0 \gg \frac{k}{2} \paren{\xi^2 + \eta^2}   \quad \forall\, (\xi, \eta) \in \Omega$, 
then $e^{\frac{ik}{2z_0} \paren{\xi^2 + \eta^2}} \approx 1$ and hence we obtain:
\begin{empheq}[innerbox=\fbox,right= \; .\quad (\textbf{Fraunhofer approximation---forward})]{align*}  \label{eq:fraunhofer_approx}
    U(x, y, z_0) \approx \frac{e^{ikz_0}}{i \lambda z_0} e^{\frac{ik}{2z_0} \paren{x^2 + y^2}}\iint U(\xi, \eta, 0) e^{-\frac{ik}{z_0}(x\xi + y \eta)}  \; d \xi d\eta
\end{empheq} 
By assuming $\paren{\lambda f_X}^2 + \paren{\lambda f_Y}^2 \ll 1$ and so $\sqrt{1 - \paren{\lambda f_X}^2 - \paren{\lambda f_Y}^2} \approx 1- \paren{\lambda f_X}^2/2 - \paren{\lambda f_Y}^2/2$, we can approximate \cref{eq:RS_fourier} as 
\begin{empheq}[innerbox=\fbox,right= \; .\quad (\textbf{Fresnel approximation---backward})]{align*}  \label{eq:fresnel_approx_back}
    U(x, y, z_0) & \approx e^{i k z_0} \iint \wh{U}\paren{f_X, f_Y, 0} e^{-i\pi z_0 \lambda(f_X^2 + f_Y^2)} e^{i 2\pi \paren{f_X x + f_Y y}}\; df_X df_Y
\end{empheq} 
This is equivalent to the forward form of Fresnel approximation, due to 
\begin{align} 
 \mc F \paren{\frac{1}{i\lambda z_0} e^{\frac{ik}{2z_0}\paren{x^2 + y^2}}} = e^{-i \pi\lambda z_0(f_X^2 + f_Y^2)}, 
\end{align}
and the convolution theorem again. 

So when we measure the field intensity on the image plane, 
\begin{align} 
  \abs{U(x, y, z_0)}^2 
  & \propto \abs{\iint \brac{U(\xi, \eta, 0) e^{\frac{ik}{2z_0} \paren{\xi^2 + \eta^2}}} e^{-\frac{ik}{z_0}(x\xi + y \eta)}  \; d \xi d\eta}^2 \\
  & \propto \abs{\iint \wh{U}\paren{f_X, f_Y, 0} e^{-i\pi z_0 \lambda(f_X^2 + f_Y^2)} e^{i 2\pi \paren{f_X x + f_Y y}}\; df_X df_Y}^2
\end{align} 
according to the Fresnel (near-field) approximation, and 
\begin{align} 
    \abs{U(x, y, z_0)}^2 \propto \abs{\iint U(\xi, \eta, 0) e^{-\frac{ik}{z_0}(x\xi + y \eta)}  \; d \xi d\eta}^2
\end{align} 
by the Fraunhofer (far-field) approximation. Detailed discussion of discretization and computation can be found in \cite[Chapter 5]{Goodman2017Introduction}.

\end{document}